\documentclass{article}
\PassOptionsToPackage{table}{xcolor}
\usepackage{iclr2027_conference,times}
\usepackage[T1]{fontenc}
\usepackage{inconsolata}
\usepackage{amsmath,amssymb,booktabs,graphicx,xcolor}
\usepackage{hyperref,url,placeins,wrapfig}
\definecolor{modelband}{HTML}{EAF1F7}
\definecolor{methodband}{HTML}{F3F3F3}
\definecolor{paperlink}{HTML}{17365D}
\hypersetup{colorlinks=true,linkcolor=paperlink,citecolor=paperlink,urlcolor=paperlink}
\DeclareMathOperator*{\argmax}{arg\,max}
\newcommand{\lpd}{LoOPSD$^{\dagger}$}
\newcommand{\sg}{\operatorname{sg}}
\title{LoOPSD: On-Policy Self-Distillation with Looped Teachers}
\iclrfinalcopy
\author{%
\parbox[t]{\dimexpr\textwidth-2\tabcolsep\relax}{\raggedright\bfseries
Wei Fan$^{1*\dagger}$, Qingyue Yang$^{3\dagger}$, Chao-Hong Tan$^{2}$,
Qian Chen$^{2}$, Wen Wang$^{2}$, Xiangang Li$^{2}$,
Kejiang Chen$^{1\ddagger}$, Weiming Zhang$^{1}$, Nenghai Yu$^{1}$}\\
\normalfont $^{1}$Anhui Province Key Laboratory of Digital Security \\
\normalfont $^{2}$Token Foundry, Alibaba Group \\
\normalfont $^{3}$Independent Researcher \\
\footnotesize\texttt{range@mail.ustc.edu.cn; \{chenkj, zhangwm, ynh\}@ustc.edu.cn} \\
\footnotesize\texttt{\{tanchaohong.ch, tanqing.cq, w.wang, lixiangang.lxg\}@alibaba-inc.com}
}
\hypersetup{
  pdftitle={LoOPSD: On-Policy Self-Distillation with Looped Teachers},
  pdfauthor={Wei Fan, Qingyue Yang, Chao-Hong Tan, Qian Chen, Wen Wang, Xiangang Li, Kejiang Chen, Weiming Zhang, Nenghai Yu}
}

\begin{document}
\maketitle
\fancyhead{}
\lhead{Preprint}
\begingroup
\renewcommand{\thefootnote}{\fnsymbol{footnote}}
\footnotetext[1]{Work done during Wei Fan's internship at Token Foundry, Alibaba Group.}
\footnotetext[2]{Equal contribution.}
\footnotetext[3]{Corresponding author.}
\endgroup
\begin{abstract}
On-policy self-distillation improves a model using guidance from the model itself, but it still requires a teacher that provides more useful predictions than the student. Existing approaches obtain such guidance by giving the teacher additional context or instructions. We ask whether it can instead come from changing how the same model uses its own layers. To address this, we introduce LoOPSD, which searches over a model's layer execution order to construct a stronger teacher and distills its advantage into the student. Specifically, we keep the model weights unchanged while searching for layers to repeat, and use the resulting teacher's predictions on student-generated prefixes to guide learning. We then alternate search and distillation, rebuilding the teacher from the updated student in each round so that the student progressively absorbs the gains from alternative layer execution. Experiments on mathematical and logical reasoning tasks show improvements across models. On GSM8K with a 400-token thinking budget, LoOPSD raises Qwen3-1.7B accuracy from 42.30\% to 79.61\%; subsequent distillation rounds yield further gains beyond the first round. Together, these findings show that by searching layer execution to construct teachers, we can turn a model's additional computation into improved student capabilities, opening a new avenue for self-distillation to derive supervision from the model's own computation.
\end{abstract}
\section{Introduction}
On-policy distillation has emerged as a promising approach to improving language models by learning from teacher predictions on student-generated trajectories \citep{agarwal2024onpolicy}. Its token-level guidance lets a student learn along the response prefixes it actually produces. Within this framework, self-distillation offers the possibility of improving reasoning without requiring a stronger external teacher: the model itself supplies the predictions used for learning \citep{zhao2026selfdistilled,sang2026crisp}. The key question is where this self-teacher obtains predictions that are more useful than those of the student.

Existing self-distillation approaches obtain such guidance by giving the teacher additional context or instructions. Additional context can provide a teacher with information beyond the student's input, while a concise instruction can elicit a different reasoning behavior \citep{zhao2026selfdistilled,sang2026crisp}. We investigate whether the same model can become a stronger teacher by running selected layers more than once, without changing their weights. Can learning from this teacher improve a student that uses each layer once in the original order?

\begin{figure}[!htbp]
\centering
\includegraphics[width=0.83\textwidth]{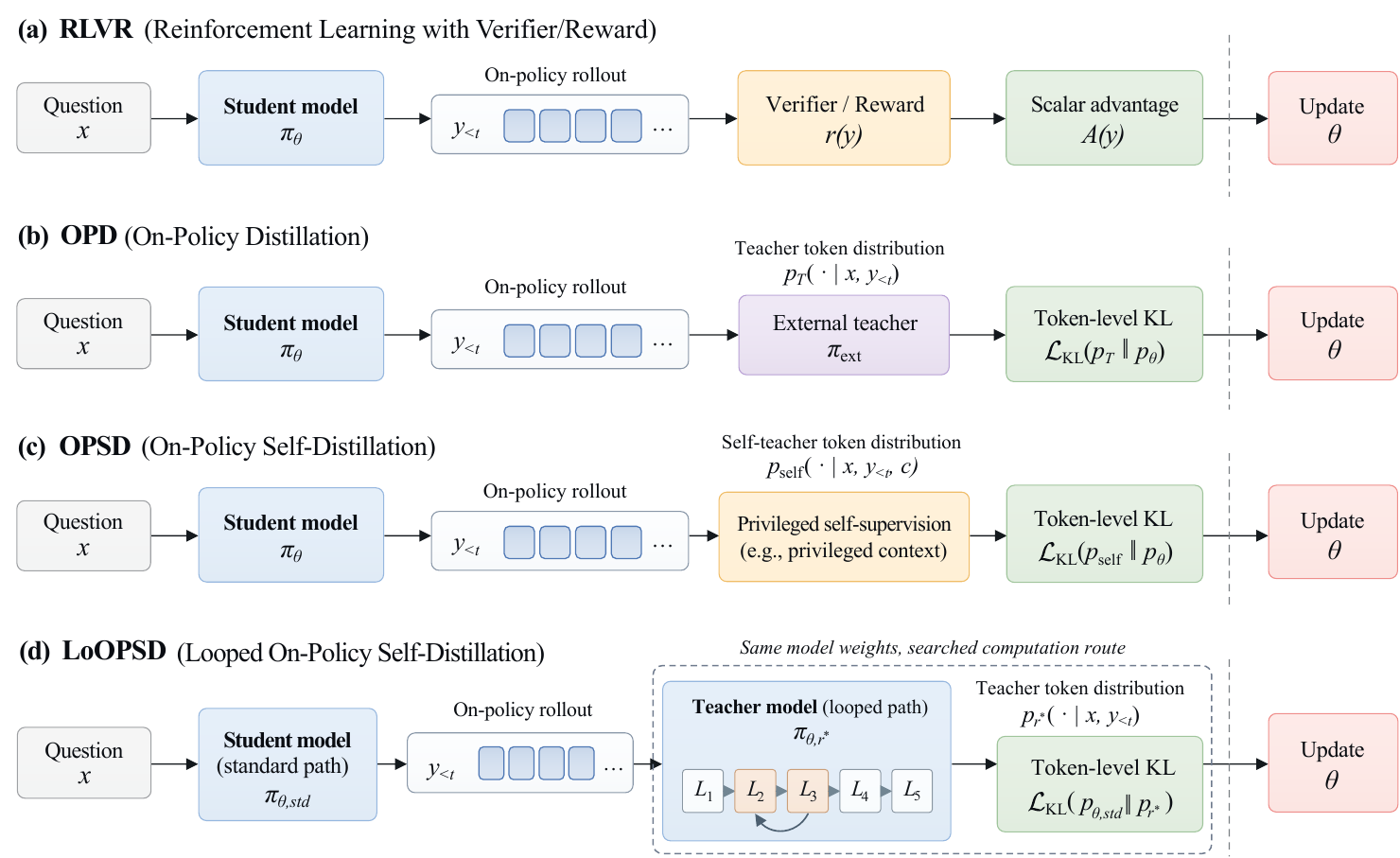}
\caption{Sources of supervision for reasoning models. LoOPSD constructs a teacher from the student's starting weights by selecting an alternative layer route with labeled examples. The resulting predictions provide supervision without an external teacher model.}
\label{fig:sources}
\end{figure}

Layer-execution research provides a concrete opportunity. Repeating or rearranging pretrained layers can improve a model's reasoning without changing its weights \citep{batorski2026macro,li2026polar}. We search for which block of layers to repeat and use the resulting model as a teacher. The research question is whether the advantage obtained through this execution can become an improvement in the student's parameters. Route search identifies how to construct the teacher; distillation determines how the student learns from its guidance.

We introduce \textbf{LoOPSD}, an on-policy self-distillation method that uses a looped execution of the current model as the teacher. We first search for a repeated layer interval using labeled questions and validate the selected route against standard execution. We then freeze the resulting teacher and train a student initialized from the same checkpoint. The student generates responses, and the teacher scores those response prefixes to provide a token-level learning signal. In this way, the teacher's additional computation guides the trajectories on which the student learns.

We alternate layer search and distillation over multiple rounds. After each round, we search for a block of layers to repeat in the updated student and use the resulting looped model as the next teacher. The student progressively absorbs the gains available from its teachers, and the process ends when validation no longer identifies a useful looped teacher or the round limit is reached.

LoOPSD also allows different prompts for the teacher and student. We evaluate this extension by instructing the teacher to reason concisely during both route search and distillation, while the student retains its original prompt.

Across mathematical and logical reasoning tasks, we find that LoOPSD improves student accuracy across model sizes and families. On GSM8K, LoOPSD improves Qwen3-1.7B from 42.30\% to 79.61\% under a 400-token thinking budget; it also improves Phi-4-reasoning-plus from 89.54\% to 94.92\%. We then examine how these gains develop through iteration, how they appear across thinking budgets and realized generation lengths, and where answer readout changes inside the student. A label-efficiency study measures the feedback needed to construct useful teachers. Together, these experiments investigate how searched layer execution supplies supervision and how students learn from it.

\section{Related Work}
\paragraph{On-policy distillation and self-distillation.}
Knowledge distillation uses teacher predictions as supervision \citep{hinton2015distilling}; Born Again Networks show that students of the same architecture can improve over their teachers \citep{furlanello2018born}. For reasoning, STaR and ReST$^{\mathrm{EM}}$ iterate generation and training on responses selected using correctness feedback \citep{zelikman2022star,singh2024rest}. On-policy distillation instead supplies teacher predictions along student-generated trajectories: GKD trains the student on its own generated responses to reduce the mismatch between training and inference \citep{agarwal2024onpolicy}, while MiniLLM develops reverse-KL distillation with on-policy optimization \citep{gu2024minillm}. Recent self-distillation methods give the teacher additional information or instructions that the student does not receive, such as reference reasoning traces in Self-Distilled Reasoner and an instruction to reason concisely in CRISP \citep{zhao2026selfdistilled,sang2026crisp}. We construct and renew teachers by searching layer execution, and allow the teacher to receive a different prompt from the student.

\paragraph{Layer search and adaptive execution.}
Universal Transformers and recurrent-depth language models are trained to apply the same layer block repeatedly, refining hidden states with shared parameters at each repetition \citep{dehghani2019universal,geiping2025recurrent}. Existing pretrained models also permit execution changes: RYS searches repeated contiguous layer blocks \citep{ng2026rys}, whereas Recirculation introduces recurrent activation connections at inference time \citep{mozer2026recirculation}. MACRO searches routes involving repeated, skipped, or combined layers \citep{batorski2026macro}, and Program-of-Layers learns which pretrained layers to execute, skip, or repeat for each input \citep{li2026polar}. These studies show that changing layer execution can improve a model's reasoning performance. LoOPSD turns this execution advantage into training supervision: it constructs a teacher through layer search and distills its guidance into a standard-path student, repeating this process as the student's weights change.

\paragraph{Reasoning under generation budgets.}
Reasoning performance often depends on how much computation a model is allowed to use during generation \citep{wei2022chain}. Test-time methods such as s1 directly control reasoning length through budget forcing \citep{muennighoff2025s1}, while training-based approaches encourage effective reasoning under constrained generation through length schedules, reward design, or compression, as in ThinkPrune, ShorterBetter, and TokenSkip \citep{hou2025thinkprune,yi2025shorterbetter,xia2025tokenskip}. Elastic Reasoning trains models explicitly for bounded generation with group relative policy optimization \citep{xu2026elastic,shao2024deepseekmath}, whereas CRISP distills guidance from a concise self-teacher \citep{sang2026crisp}. These methods primarily control or improve the use of generated reasoning tokens. LoOPSD instead derives supervision from additional computation in layer execution and transfers this advantage to a standard-execution student. We therefore study whether execution-derived supervision improves both reasoning accuracy and realized token use under fixed thinking budgets, and whether it can be combined with concise teacher prompting.

\section{LoOPSD}
\label{sec:method}
LoOPSD turns a model's execution flexibility into training supervision. A round has three components: selecting a useful route, learning from the resulting frozen teacher on student-generated prefixes, and constructing the next teacher from the updated checkpoint.

\begin{figure}[!htbp]
\centering
\includegraphics[width=\textwidth]{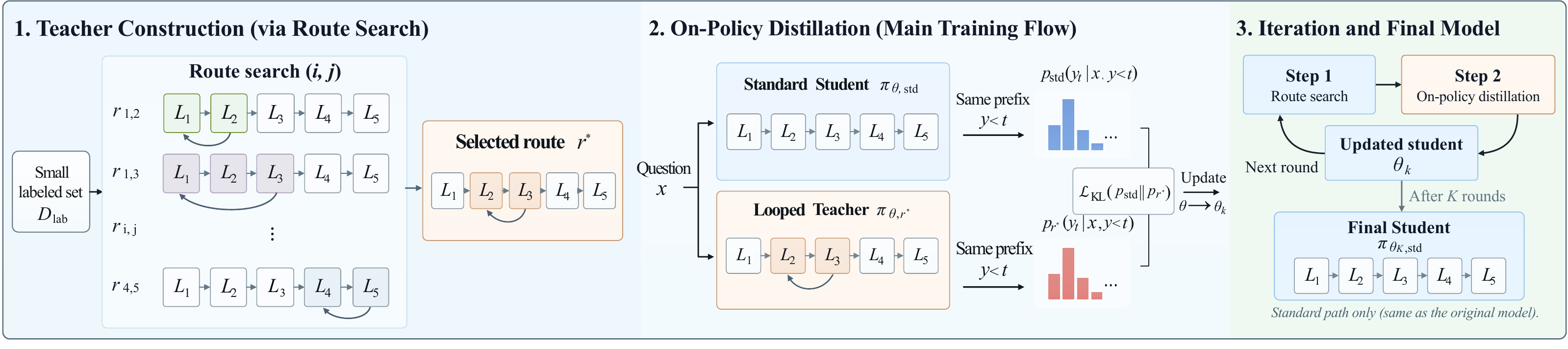}
\caption{\textbf{Overview of LoOPSD.} \textbf{(1) Teacher construction:} Using a small labeled set, we try different choices of layers to repeat, then select the best-performing route to define the teacher. \textbf{(2) On-policy distillation:} The student generates responses using the original layer order. The frozen teacher evaluates the same response prefixes and provides token-level supervision to update the student. \textbf{(3) Iteration and final model:} The updated student is used to build the teacher for the next round. After $K$ rounds, the final student uses the original layer order.}
\label{fig:method}
\end{figure}

\subsection{Selecting the teacher's computation}
Let $p_{\theta,\mathrm{std}}$ denote a model with parameters $\theta$ that runs layers $1,\ldots,L$ in their original order. A route $r$ specifies which layers are run and in what order. In our main experiments, a block of consecutive layers is run twice, reusing the same weights \citep{ng2026rys}. We write $p_{\theta,r}$ for the model's output distribution under route $r$.

Repeating layers $i$ through $j$ ($1\leq i\leq j\leq L$) gives the route
\begin{equation}
 r_{i,j}=(1,\ldots,i-1,\;i,\ldots,j,\;i,\ldots,j,\;j+1,\ldots,L).
 \label{eq:route-family}
\end{equation}
The teacher performs $L+(j-i+1)$ layer evaluations per token. Each repeated layer uses the same pretrained parameters on the hidden state produced by the previous execution. The route is shared across questions within a round. Selecting it changes the teacher's predictions without requiring a larger parameter checkpoint.

At round $k$, we evaluate candidate routes on labeled search questions in stages, retaining the best-performing routes in $\mathcal C_k$. We then compare these candidates with the standard route on a separate validation set $\mathcal D_{\mathrm{val}}$, using the same prompt and token budgets:
\begin{equation}
 r_k^*=\argmax_{r\in\mathcal C_k\cup\{\mathrm{std}\}}
 \frac{1}{|\mathcal D_{\mathrm{val}}|}\sum_{(x,a)\in\mathcal D_{\mathrm{val}}}
 V\!\left(\operatorname{Decode}_{B,A}(p_{\theta_k,r},x),a\right).
 \label{eq:route-selection}
\end{equation}
Here, $V$ scores a generated answer against its reference answer, and $B$ and $A$ are the maximum numbers of thinking and answer tokens, respectively. If the selected route scores higher than the standard route on the validation set, it defines the teacher $q_k=p_{\theta_k,r_k^*}$. The teacher's weights and route remain fixed during the following distillation round. Candidate-screening details are provided in Appendix~\ref{sec:implementation}.

\subsection{On-Policy Distillation}
Given the selected teacher, the student generates responses from its current policy, and the teacher evaluates the same student-generated prefixes. We initialize the standard-path student from $\theta_k$ and sample responses $y$ for questions $x$ from $\mathcal D_q$. We train the student to match the teacher's next-token distribution using the student-to-teacher reverse Kullback--Leibler (KL) divergence \citep{gu2024minillm}:
\begin{equation}
 \mathcal L(\theta)=
 \mathbb E_{x\sim\mathcal D_q,\,y\sim p_{\mathrm{old}}(\cdot\mid x)}
 \left[\frac{1}{T}\sum_{t=1}^{T}
 D_{\mathrm{KL}}\!\left(
 p_{\theta,\mathrm{std}}(\cdot\mid x,y_{<t})\,
 \Vert\,q_k(\cdot\mid x,y_{<t})\right)\right],
 \label{eq:ideal-kl}
\end{equation}
Here, $p_{\mathrm{old}}$ is the student policy used to generate the responses. The teacher remains fixed throughout the round, and reference answers are used only for teacher selection rather than distillation. We optimize this reverse-KL objective using the sampled-token on-policy distillation implementation described in Appendix~\ref{sec:implementation}. Every response prefix used for distillation is generated by the student, allowing the teacher to provide local guidance along the trajectories encountered by the student.

\subsection{Iterative Self-Distillation and Extensions}
\label{sec:iterative-extensions}
\paragraph{Alternating search and distillation.}
The two steps above can be repeated in alternation to improve the model over multiple rounds. Search constructs a looped teacher from the current model, and distillation transfers its guidance into a standard-path student. The resulting student becomes the starting point for the next search, allowing subsequent rounds to build on what the model has already learned. We continue this process until validation no longer identifies a useful loop or a preset round limit is reached.

\paragraph{Extension: Different Prompts for Teacher and Student.}
So far, the teacher and student have received the same prompt. LoOPSD also allows the teacher to receive additional instructions or information while the student retains its original prompt. We search for a repeated-layer interval using the teacher's prompt, then distill the resulting teacher's predictions on student-generated prefixes. The search procedure and distillation objective remain unchanged. In our experiments, we instruct the teacher to reason concisely and denote this configuration by \lpd{}.

\section{Experimental Setup}
\label{sec:setup}
\paragraph{Models and tasks.}
We evaluate Qwen3-1.7B, 4B, and 8B \citep{yang2025qwen3}, and DeepSeek-R1-Distill-Llama-8B \citep{deepseekai2025r1}; Phi-4-reasoning-plus \citep{abdin2025phi4reasoning} and Qwen3-14B provide larger-model comparisons. We evaluate mathematical reasoning on GSM8K \citep{cobbe2021gsm8k}, SVAMP \citep{patel2021svamp}, and MATH-500 \citep{hendrycks2021math,lightman2024verify}, and logical reasoning on ProofWriter \citep{tafjord2021proofwriter}. We additionally evaluate GSM8K-trained students on GSM-Plus \citep{li2024gsmplus} as a transfer benchmark, without further target-task training.

\paragraph{Data splits.}
For the task-specific experiments, we train a separate student for each task. For MATH-500, training uses the PRM800K split \citep{lightman2024verify}. For LoOPSD on GSM8K, we hold out 100 training questions for validation and use the remaining 7,373 questions for distillation. Route search uses labeled answers from a subset of 1,000 questions in this distillation pool. Reference answers are used for teacher selection, but are not provided during distillation. Test questions are excluded from route search and checkpoint selection.

\paragraph{Training.}
We use the repeated-layer route defined in Section 3.1: a contiguous interval of pretrained layers is executed twice with shared weights. Each LoOPSD round trains the student for 200 updates against a frozen teacher, with at most 1,024 generated tokens per training response. We allow up to ten rounds, following the validation-based continuation rule in Section~\ref{sec:iterative-extensions}. For \lpd{}, the different-prompt extension introduced in Section~\ref{sec:iterative-extensions}, the teacher receives a concise reasoning instruction during both route search and distillation, while the student retains its original prompt. Appendix~\ref{sec:implementation} provides the route-search procedure, optimization, and continuation details.

\paragraph{Baselines.}
We compare LoOPSD and \lpd{} with the original model, Elastic Reasoning \citep{xu2026elastic}, and CRISP \citep{sang2026crisp}. Elastic Reasoning uses GRPO with a limit on thinking tokens during training. For CRISP, we use a concise standard-path teacher for one 200-update round with the same distillation implementation as LoOPSD.

\paragraph{Evaluation.}
By default, we use greedy decoding with up to 400 thinking tokens followed by up to 128 answer tokens. We report accuracy using task-specific rules to compare generated answers with reference answers. All students use the original prompt and standard layer order. For the budget experiments, we either obtain a separate student for each target search budget or vary only the test-time thinking budget of a fixed student. The figure captions specify which setting is used.

\section{Results}
\subsection{Improved Student Reasoning}
\label{sec:main-results}
\paragraph{Reasoning across models and tasks.}
To assess learning gains from looped teachers, we first compare R1 (Round 1, 200 student updates) students in Table~\ref{tab:main}. At R1, \lpd{} already outperforms CRISP (also 200 student updates) in most evaluated settings. LoOPSD also improves the original models in most settings. Further search-and-distillation rounds generally improve accuracy over R1. The resulting final students (Final) surpass the original models in all evaluated pairs. Gains span mathematical and logical reasoning and transfer to GSM-Plus without target-task training. These results support looped-teacher supervision as a source of iterative student improvement.

\begin{table}[!htbp]
\centering
\caption{Accuracy (\%) across four models and five benchmarks, using greedy decoding with 400 thinking and 128 answer tokens. For LoOPSD variants, entries report first-round (R1) $\to$ final accuracy. R1 denotes the student after the first distillation round; final denotes the checkpoint obtained by the iterative continuation procedure. GSM-Plus evaluates the corresponding GSM8K-trained checkpoints.}
\label{tab:main}
\small
\setlength{\tabcolsep}{3.5pt}
\renewcommand{\arraystretch}{1.01}
\begin{tabular}{p{94pt}*{5}{>{\raggedleft\arraybackslash}p{\dimexpr(\linewidth-136pt)/5\relax}}}
\toprule
\textbf{Method} & \multicolumn{1}{c}{\textbf{GSM8K}} & \multicolumn{1}{c}{\textbf{GSM-Plus}} & \multicolumn{1}{c}{\textbf{SVAMP}} & \multicolumn{1}{c}{\textbf{MATH-500}} & \multicolumn{1}{c}{\textbf{ProofWriter}} \\
\specialrule{0.5pt}{3pt}{0pt}
\rowcolor{modelband}
\multicolumn{6}{l}{\bfseries\fontsize{8.5}{9.5}\selectfont Qwen3-1.7B} \\
Original & 42.30 & 31.70 & 71.33 & 33.20 & 42.50 \\
Elastic Reasoning & 46.25 & 30.20 & 73.00 & 34.00 & 45.00 \\
CRISP & 76.65 & 58.50 & 88.33 & 37.80 & 43.40 \\
\rowcolor{methodband}
LoOPSD \textcolor{black!55}{(R1$\to$Final)} & \mbox{\makebox[21pt][r]{78.09}\makebox[9pt][c]{$\to$}\makebox[21pt][r]{79.61}} & \mbox{\makebox[21pt][r]{56.50}\makebox[9pt][c]{$\to$}\makebox[21pt][r]{61.40}} & \mbox{\makebox[21pt][r]{87.67}\makebox[9pt][c]{$\to$}\makebox[21pt][r]{87.67}} & \mbox{\makebox[21pt][r]{44.40}\makebox[9pt][c]{$\to$}\makebox[21pt][r]{44.80}} & \mbox{\makebox[21pt][r]{42.10}\makebox[9pt][c]{$\to$}\makebox[21pt][r]{\textbf{56.20}}} \\
\rowcolor{methodband}
LoOPSD$^{\dagger}$ \textcolor{black!55}{(R1$\to$Final)} & \mbox{\makebox[21pt][r]{78.32}\makebox[9pt][c]{$\to$}\makebox[21pt][r]{\textbf{79.68}}} & \mbox{\makebox[21pt][r]{59.30}\makebox[9pt][c]{$\to$}\makebox[21pt][r]{\textbf{61.80}}} & \mbox{\makebox[21pt][r]{88.67}\makebox[9pt][c]{$\to$}\makebox[21pt][r]{\textbf{91.67}}} & \mbox{\makebox[21pt][r]{37.20}\makebox[9pt][c]{$\to$}\makebox[21pt][r]{\textbf{48.20}}} & \mbox{\makebox[21pt][r]{46.80}\makebox[9pt][c]{$\to$}\makebox[21pt][r]{54.10}} \\
\specialrule{0.5pt}{0pt}{0pt}
\rowcolor{modelband}
\multicolumn{6}{l}{\bfseries\fontsize{8.5}{9.5}\selectfont Qwen3-4B} \\
Original & 74.98 & 56.40 & 87.33 & 36.00 & 63.30 \\
Elastic Reasoning & 75.59 & 58.90 & 89.33 & 37.40 & 67.90 \\
CRISP & 87.72 & 70.10 & \textbf{92.00} & 44.60 & 66.00 \\
\rowcolor{methodband}
LoOPSD \textcolor{black!55}{(R1$\to$Final)} & \mbox{\makebox[21pt][r]{84.38}\makebox[9pt][c]{$\to$}\makebox[21pt][r]{89.76}} & \mbox{\makebox[21pt][r]{67.00}\makebox[9pt][c]{$\to$}\makebox[21pt][r]{\textbf{74.20}}} & \mbox{\makebox[21pt][r]{90.33}\makebox[9pt][c]{$\to$}\makebox[21pt][r]{88.67}} & \mbox{\makebox[21pt][r]{46.00}\makebox[9pt][c]{$\to$}\makebox[21pt][r]{50.20}} & \mbox{\makebox[21pt][r]{68.00}\makebox[9pt][c]{$\to$}\makebox[21pt][r]{\textbf{76.40}}} \\
\rowcolor{methodband}
LoOPSD$^{\dagger}$ \textcolor{black!55}{(R1$\to$Final)} & \mbox{\makebox[21pt][r]{87.04}\makebox[9pt][c]{$\to$}\makebox[21pt][r]{\textbf{91.13}}} & \mbox{\makebox[21pt][r]{69.00}\makebox[9pt][c]{$\to$}\makebox[21pt][r]{73.20}} & \mbox{\makebox[21pt][r]{91.33}\makebox[9pt][c]{$\to$}\makebox[21pt][r]{91.33}} & \mbox{\makebox[21pt][r]{52.80}\makebox[9pt][c]{$\to$}\makebox[21pt][r]{\textbf{60.40}}} & \mbox{\makebox[21pt][r]{66.20}\makebox[9pt][c]{$\to$}\makebox[21pt][r]{71.50}} \\
\specialrule{0.5pt}{0pt}{0pt}
\rowcolor{modelband}
\multicolumn{6}{l}{\bfseries\fontsize{8.5}{9.5}\selectfont Qwen3-8B} \\
Original & 76.04 & 61.00 & 89.33 & 39.00 & 65.40 \\
Elastic Reasoning & 76.88 & 61.20 & 91.33 & 39.80 & 70.50 \\
CRISP & 89.61 & 70.00 & 92.33 & 45.80 & 65.90 \\
\rowcolor{methodband}
LoOPSD \textcolor{black!55}{(R1$\to$Final)} & \mbox{\makebox[21pt][r]{87.72}\makebox[9pt][c]{$\to$}\makebox[21pt][r]{90.67}} & \mbox{\makebox[21pt][r]{69.70}\makebox[9pt][c]{$\to$}\makebox[21pt][r]{74.20}} & \mbox{\makebox[21pt][r]{91.33}\makebox[9pt][c]{$\to$}\makebox[21pt][r]{92.33}} & \mbox{\makebox[21pt][r]{45.00}\makebox[9pt][c]{$\to$}\makebox[21pt][r]{49.60}} & \mbox{\makebox[21pt][r]{72.90}\makebox[9pt][c]{$\to$}\makebox[21pt][r]{79.40}} \\
\rowcolor{methodband}
LoOPSD$^{\dagger}$ \textcolor{black!55}{(R1$\to$Final)} & \mbox{\makebox[21pt][r]{92.49}\makebox[9pt][c]{$\to$}\makebox[21pt][r]{\textbf{92.49}}} & \mbox{\makebox[21pt][r]{74.30}\makebox[9pt][c]{$\to$}\makebox[21pt][r]{\textbf{74.30}}} & \mbox{\makebox[21pt][r]{94.00}\makebox[9pt][c]{$\to$}\makebox[21pt][r]{\textbf{94.00}}} & \mbox{\makebox[21pt][r]{54.00}\makebox[9pt][c]{$\to$}\makebox[21pt][r]{\textbf{65.80}}} & \mbox{\makebox[21pt][r]{73.10}\makebox[9pt][c]{$\to$}\makebox[21pt][r]{\textbf{81.60}}} \\
\specialrule{0.5pt}{0pt}{0pt}
\rowcolor{modelband}
\multicolumn{6}{l}{\bfseries\fontsize{8.5}{9.5}\selectfont DeepSeek-R1-Distill-Llama-8B} \\
Original & 65.05 & 49.00 & 76.67 & 38.60 & 39.20 \\
Elastic Reasoning & 73.09 & 53.80 & 87.00 & 42.20 & \textbf{51.80} \\
CRISP & 68.76 & 52.80 & 80.67 & 33.00 & 38.30 \\
\rowcolor{methodband}
LoOPSD \textcolor{black!55}{(R1$\to$Final)} & \mbox{\makebox[21pt][r]{74.45}\makebox[9pt][c]{$\to$}\makebox[21pt][r]{76.04}} & \mbox{\makebox[21pt][r]{52.20}\makebox[9pt][c]{$\to$}\makebox[21pt][r]{57.60}} & \mbox{\makebox[21pt][r]{80.67}\makebox[9pt][c]{$\to$}\makebox[21pt][r]{88.33}} & \mbox{\makebox[21pt][r]{39.00}\makebox[9pt][c]{$\to$}\makebox[21pt][r]{\textbf{43.40}}} & \mbox{\makebox[21pt][r]{49.60}\makebox[9pt][c]{$\to$}\makebox[21pt][r]{51.10}} \\
\rowcolor{methodband}
LoOPSD$^{\dagger}$ \textcolor{black!55}{(R1$\to$Final)} & \mbox{\makebox[21pt][r]{80.89}\makebox[9pt][c]{$\to$}\makebox[21pt][r]{\textbf{81.65}}} & \mbox{\makebox[21pt][r]{60.20}\makebox[9pt][c]{$\to$}\makebox[21pt][r]{\textbf{59.50}}} & \mbox{\makebox[21pt][r]{88.67}\makebox[9pt][c]{$\to$}\makebox[21pt][r]{\textbf{88.67}}} & \mbox{\makebox[21pt][r]{37.20}\makebox[9pt][c]{$\to$}\makebox[21pt][r]{\textbf{43.40}}} & \mbox{\makebox[21pt][r]{47.80}\makebox[9pt][c]{$\to$}\makebox[21pt][r]{48.90}} \\
\specialrule{0.7pt}{1pt}{0pt}
\end{tabular}
\par\vspace{4pt}
\begin{minipage}{\linewidth}\footnotesize
\textbf{R1$\to$Final}: first-round student accuracy $\to$ final student accuracy. \textbf{Bold}: best final accuracy per model and task. $\dagger$: concise teacher prompt. All students use their original prompt and standard layer order. GSM-Plus uses the fixed 1000-question subset; other columns use task-specific training.
\end{minipage}
\end{table}

\paragraph{Different Teacher Prompts.}
To evaluate the different-prompt extension in Section~\ref{sec:iterative-extensions}, we compare LoOPSD with \lpd{}, which uses a concise teacher prompt. On MATH-500 at R1, Qwen3-4B students reach 46.00\% with LoOPSD and 52.80\% with \lpd{}, versus 44.60\% with CRISP; the corresponding 8B scores are 45.00\%, 54.00\%, and 45.80\%. At Final, \lpd{} reaches 60.40\% and 65.80\%, respectively. Concise teacher prompting thus strengthens supervision throughout this iterative process.

\paragraph{Scaling to larger models.}
We evaluate Phi-4-reasoning-plus and Qwen3-14B on GSM8K at a 400-token thinking budget to test whether the gains extend to larger models (Table~\ref{tab:large-models}). The R1 students of \lpd{} reach 95.07\% and 93.63\%, respectively, versus CRISP's 92.95\% and 92.57\%. At Final, LoOPSD reaches 94.92\% and 94.39\%. These results extend the effectiveness of execution-derived supervision to the 14B scale.
\begin{table}[!htbp]
\centering
\caption{GSM8K accuracy (\%) of 14B-scale reasoning models with a 400-token thinking budget and greedy decoding. LoOPSD entries report first-round (R1) $\to$ final accuracy. All students use standard execution and their original prompt. LoOPSD achieves the highest final accuracy for both model families.}
\label{tab:large-models}
\small
\setlength{\tabcolsep}{12pt}
\renewcommand{\arraystretch}{1.12}
\resizebox{0.90\width}{!}{%
\begin{tabular}{lrr}
\toprule
\rowcolor{modelband}
Method & Phi-4-reasoning-plus & Qwen3-14B \\
\midrule
Original & 89.54 & 83.40 \\
Elastic Reasoning & 89.69 & 85.14 \\
CRISP & 92.95 & 92.57 \\
\rowcolor{methodband}
LoOPSD \textcolor{black!55}{(R1$\to$Final)} & \mbox{91.89\kern1pt$\to$\kern1pt\textbf{94.92}} & \mbox{90.07\kern1pt$\to$\kern1pt\textbf{94.39}} \\
\rowcolor{methodband}
LoOPSD$^{\dagger}$ \textcolor{black!55}{(R1$\to$Final)} & \mbox{95.07\kern1pt$\to$\kern1pt93.18} & \mbox{93.63\kern1pt$\to$\kern1pt93.63} \\
\bottomrule
\end{tabular}%
}
\par\vspace{3pt}
\begin{minipage}{0.94\linewidth}\footnotesize
\textbf{R1$\to$Final}: first-round student accuracy $\to$ final student accuracy. $\dagger$: concise teacher prompt; all students use the original prompt. The Qwen3-14B Elastic evaluation has a paired original score of 83.85\%.
\end{minipage}
\end{table}

\subsection{Reasoning Quality under Limited Thinking Budgets}
\label{sec:budgets}
\begin{figure}[!t]
\centering
\includegraphics[width=\textwidth]{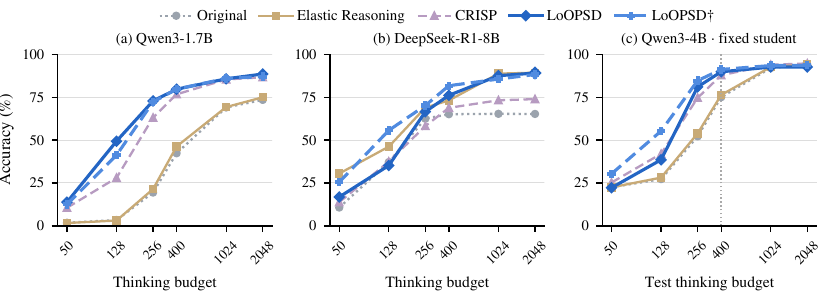}
\caption{Accuracy across thinking budgets on GSM8K. (a,b) Qwen3-1.7B and DeepSeek-R1-Distill-Llama-8B students obtained separately for each target search budget and evaluated at that budget. (c) Qwen3-4B students obtained with a 400-token search budget, with each checkpoint fixed across test budgets. LoOPSD improves reasoning across a range of budgets, and fixed students adapt to different test budgets.}
\label{fig:budgets}
\label{fig:fixed400}
\end{figure}

\paragraph{Effectiveness across thinking budgets.}
Figure~\ref{fig:budgets}(a,b) evaluates Qwen3-1.7B and DeepSeek-R1-Distill-Llama-8B on GSM8K across six thinking budgets. We obtain a student for each target search budget and evaluate it at that budget. LoOPSD improves accuracy over the original model across a broad range of budgets, and its concise-teacher extension improves both models at every tested budget. These results show that the benefits extend across different budgets for reasoning, rather than being confined to one budget. Appendix~\ref{sec:extended} provides the full comparisons.

\paragraph{Adaptation to different test budgets.}
Figure~\ref{fig:fixed400}(c) evaluates whether a learned student can also accommodate budgets other than the one used for route search. We fix the Qwen3-4B students obtained with a 400-token search budget and vary only their test budgets. The students achieve higher or broadly comparable accuracy to the original model across the tested range, with substantial gains at several budgets outside the search configuration. Thus, a single learned student can support different thinking budgets without further search or training.

\subsection{Token Efficiency of Learned Students}
\label{sec:length}
\begin{figure}[!htbp]
\centering
\makebox[\textwidth][c]{\includegraphics[width=0.92\textwidth]{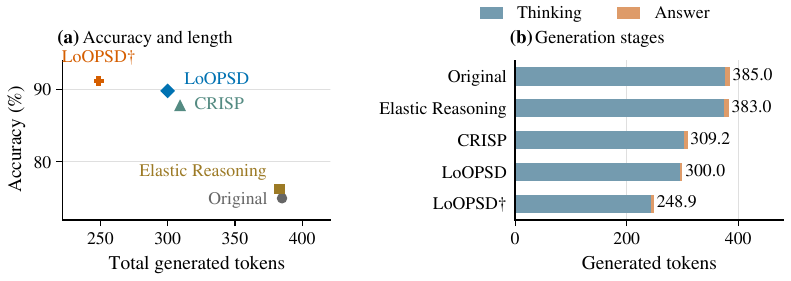}}
\caption{Accuracy and generation length of different methods on GSM8K with Qwen3-4B. All models use a 400-token thinking budget. (a) Accuracy versus average total generated tokens. (b) Average thinking and answer lengths. LoOPSD improves accuracy while shortening generation; LoOPSD$^{\dagger}$ achieves the highest accuracy with the fewest thinking tokens.}
\label{fig:length}
\end{figure}

We compare accuracy and actual generation length under a shared thinking budget to assess the token efficiency of LoOPSD students. Figure~\ref{fig:length} evaluates Qwen3-4B on GSM8K at a 400-token thinking budget, measuring accuracy and generation length from the same responses.

LoOPSD produces more accurate answers with fewer generated tokens: accuracy rises from 74.98\% to 89.76\%, while average total generation falls from 385.04 to 299.98 tokens. Its concise-teacher extension further improves accuracy while shortening generation. The length breakdown in Figure~\ref{fig:length}(b) shows shorter thinking sequences alongside shorter total responses. Together, these results show that LoOPSD students achieve higher accuracy with shorter generated responses under the same thinking budget.

\section{Analysis}
\label{sec:analysis}
\subsection{Label Efficiency}
We compare methods under matched labeled-data budgets to assess the label efficiency of LoOPSD. We evaluate LoOPSD and GRPO with truncated thinking across different label budgets while matching training updates. This comparison examines two ways of using labeled feedback: LoOPSD evaluates alternative layer execution orders on a small labeled set and turns the selected teacher's predictions into dense token-level supervision, whereas GRPO scores multiple responses sampled from the same model and updates it using group-relative rewards.

Specifically, we use Qwen3-1.7B on GSM8K with budgets of 200, 400, and 1100 labeled examples, each including 100 validation examples. Both methods receive 200 training updates, and LoOPSD's distillation question pool remains fixed.

\begin{table}[!htbp]
\centering
\caption{Label efficiency on GSM8K with Qwen3-1.7B. LoOPSD and matched-label GRPO each receive 200 updates; label totals include 100 validation examples, and LoOPSD uses a fixed question-only distillation pool. LoOPSD already obtains most of its observed gain with 200 labels.}
\label{tab:labels}
\small
\setlength{\tabcolsep}{6pt}
\renewcommand{\arraystretch}{1.16}
\resizebox{0.88\width}{!}{%
\begin{tabular}{p{108pt}*{3}{>{\raggedleft\arraybackslash}p{\dimexpr(0.84\linewidth-156pt)/3\relax}}}
\toprule
& \multicolumn{3}{c}{Total labeled examples} \\
\cmidrule(lr){2-4}
Method & 200 & 400 & 1100 \\
\midrule
Original & \multicolumn{3}{c}{42.30 (shared baseline)} \\
Matched-label GRPO & 43.90 & 43.52 & 44.05 \\
\rowcolor{methodband}
LoOPSD & \textbf{78.01} & \textbf{78.39} & \textbf{78.62} \\
\bottomrule
\end{tabular}%
}
\par\vspace{4pt}
\begin{minipage}{0.84\linewidth}\footnotesize
Totals include 100 validation labels. LoOPSD distills on a fixed question-only pool; GRPO uses the matched labeled pool. Label counts and updates are matched; search compute and unlabeled-question access differ. Answer cap: 128 tokens.
\end{minipage}
\end{table}

LoOPSD achieves higher accuracy at every label budget, with most of the improvement already obtained from the smallest labeled pool (Table~\ref{tab:labels}). With a fixed question-only distillation pool, LoOPSD obtains most of its observed improvement using only a small number of labeled examples for teacher construction.

\begin{figure}[!t]
\centering
\includegraphics[width=0.94\textwidth]{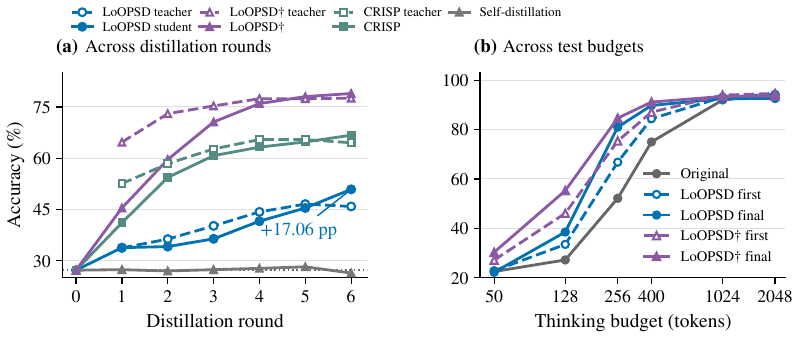}
\caption{Iterative learning on GSM8K with Qwen3-4B. (a) Each round shows the looped teacher constructed from the current model and the standard-path student obtained after distillation. We compare LoOPSD, LoOPSD$^{\dagger}$, CRISP, and standard-path self-distillation across six 200-update rounds, evaluated at a 128-token thinking budget. (b) First-round and final students from the respective LoOPSD and LoOPSD$^{\dagger}$ training trajectories, each held fixed across test budgets. Panel (a) includes matched-update standard-path controls under both the original and concise teacher prompts, while panel (b) shows the additional gains obtained across LoOPSD iterations.}
\label{fig:rounds}
\end{figure}

\subsection{Ablation Studies}
We examine two questions: whether looped teachers provide better supervision than standard-path self-teachers, and whether iterating teacher construction and distillation yields further gains beyond a single round.

\paragraph{Effect of looped teacher construction.}
To isolate the contribution of looped execution, we replace LoOPSD's looped teachers with standard-path self-teachers, keeping the prompts, distillation objective, and student-update schedule unchanged. Figure~\ref{fig:rounds}(a) evaluates all four methods on Qwen3-4B/GSM8K with a 128-token thinking budget across six 200-update rounds. LoOPSD reaches 50.87\% accuracy after 1,200 updates, compared with 26.38\% for standard-path self-distillation. We apply the same ablation to the different-prompt extension introduced in Section~\ref{sec:iterative-extensions}. For LoOPSD$^{\dagger}$, we retain the concise teacher prompt and use multi-round CRISP as the standard-path control under the same student-update schedule. LoOPSD$^{\dagger}$ also outperforms this control. Across both teacher-prompt settings, the looped-teacher variants produce stronger standard-path students than the corresponding standard-path self-distillation trajectories in this experiment.

\paragraph{Effect of iterative LoOPSD rounds.}
To assess the benefit of continuing beyond one round, we compare stopping after the first distillation round with continuing the full search-and-distillation procedure. Along the same training run, student accuracy increases from 33.81\% after the first round to 50.87\% after the sixth, a gain of 17.06 percentage points (Figure~\ref{fig:rounds}(a)). These results support the iterative design of LoOPSD: repeating teacher construction and distillation improves the standard-path student beyond what is achieved by a single round.

\subsection{Layerwise Answer Readout after Distillation}
\label{sec:readout}
\begin{wrapfigure}{r}{0.51\textwidth}
\centering
\includegraphics[width=\linewidth]{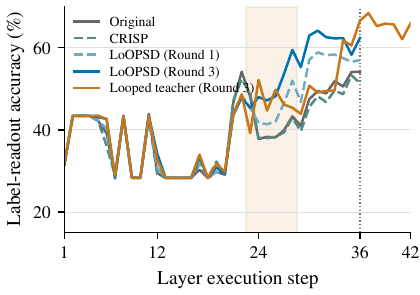}
\caption{Layerwise answer readout on ProofWriter with Qwen3-4B under shared contexts and a fixed output projection. Shading marks repeated teacher execution. Student gains concentrate in later layers.}
\label{fig:readout}
\end{wrapfigure}
To examine how distillation changes the standard student, we compare layerwise answer readout using the logit lens \citep{nostalgebraist2020logitlens} on identical ProofWriter contexts with the original model's readout head. Keeping both the input text and readout fixed allows us to compare the models without differences in the thinking sequences they generate.

Figure~\ref{fig:readout} shows that readout accuracies remain close in the early layers, while the student gains become clear in the later layers. This improvement is more pronounced after the third round than after the first, showing how late-layer answer readout develops over successive rounds. The looped teacher reaches a higher final readout accuracy with additional layer executions, whereas the student obtains its improvements within the original layer sequence. Notably, these changes arise without directly matching intermediate representations: the distillation objective supervises only output-layer predictions. Learning from the looped teacher therefore also strengthens answer readout from the standard student's intermediate layers.

\par
\ifnum\value{WF@wrappedlines}>1
\vspace{\dimexpr\baselineskip*\value{WF@wrappedlines}\relax}
\fi
\WFclear

\section{Conclusion}
We introduced LoOPSD, which searches layer execution to construct self-teachers and distills their predictions on student-generated trajectories. By alternating search and learning, we obtain further student gains beyond the first round. Across mathematical and logical reasoning tasks, the resulting students improve in accuracy, while budget, generation-length, and layerwise analyses characterize the learned changes. LoOPSD also supports different teacher prompts, as demonstrated by our concise-teacher extension. These findings show that we can turn a model's additional computation into improved student capabilities, opening a new avenue for self-distillation to derive supervision from the model's own computation.

\FloatBarrier
\section*{Reproducibility Statement}
We provide the LoOPSD procedure in Section~\ref{sec:method}, including route selection, on-policy distillation, and iterative teacher construction. Section~\ref{sec:setup} specifies the evaluated models, tasks, data splits, reasoning budgets, baselines, and evaluation protocol. Appendix~\ref{sec:implementation} provides optimization, generation, route-search, continuation, and scoring details. Additional appendices provide supporting results and analyses of teacher construction, transfer, reasoning budgets, and layerwise behavior. Test examples are excluded from route search and checkpoint selection.

\section*{AI Use Statement}
In this work, we used generative AI tools to provide feedback on experimental design and methodology, assist with result interpretation, and support parts of the implementation. We also used generative AI tools for literature search, language editing, manuscript organization, drafting and revising text, and creating or refining figures. We did not use generative AI tools to generate synthetic datasets or formulate or prove mathematical claims. All AI-assisted code was reviewed and tested by the authors, and AI-assisted literature suggestions, analyses, and manuscript content were checked against the relevant code, experiments, and source literature. We take responsibility for the final content of this work, including all text, claims, code, and artifacts produced with the aid of generative AI.

\bibliography{references}

\begin{thebibliography}{32}
\providecommand{\natexlab}[1]{#1}
\providecommand{\url}[1]{\texttt{#1}}
\expandafter\ifx\csname urlstyle\endcsname\relax
  \providecommand{\doi}[1]{doi: #1}\else
  \providecommand{\doi}{doi: \begingroup \urlstyle{rm}\Url}\fi

\bibitem[Abdin et~al.(2025)Abdin, Agarwal, Awadallah, Balachandran, Behl, Chen,
  de~Rosa, Gunasekar, Javaheripi, Joshi, Kauffmann, Lara, Mendes, Mitra, Nushi,
  Papailiopoulos, Saarikivi, Shah, Shrivastava, Vineet, Wu, Yousefi, and
  Zheng]{abdin2025phi4reasoning}
Marah Abdin, Sahaj Agarwal, Ahmed Awadallah, Vidhisha Balachandran, Harkirat
  Behl, Lingjiao Chen, Gustavo de~Rosa, Suriya Gunasekar, Mojan Javaheripi,
  Neel Joshi, Piero Kauffmann, Yash Lara, Caio C{\'e}sar~Teodoro Mendes,
  Arindam Mitra, Besmira Nushi, Dimitris Papailiopoulos, Olli Saarikivi, Shital
  Shah, Vaishnavi Shrivastava, Vibhav Vineet, Yue Wu, Safoora Yousefi, and
  Guoqing Zheng.
\newblock Phi-4-reasoning technical report, 2025.
\newblock URL \url{https://arxiv.org/abs/2504.21318}.

\bibitem[Agarwal et~al.(2024)Agarwal, Vieillard, Zhou, Stanczyk, Ramos, Geist,
  and Bachem]{agarwal2024onpolicy}
Rishabh Agarwal, Nino Vieillard, Yongchao Zhou, Piotr Stanczyk, Sabela Ramos,
  Matthieu Geist, and Olivier Bachem.
\newblock On-policy distillation of language models: Learning from
  self-generated mistakes.
\newblock In \emph{International Conference on Learning Representations}, 2024.
\newblock URL \url{https://openreview.net/forum?id=3zKtaqxLhW}.

\bibitem[Batorski et~al.(2026)Batorski, Pourhadi, Aitaza, Spurek, and
  Swoboda]{batorski2026macro}
Pawe{\l} Batorski, Abtin Pourhadi, Akylgali Aitaza, Przemys{\l}aw Spurek, and
  Paul Swoboda.
\newblock {MACRO}: Markov chain routing of transformer layers, 2026.
\newblock URL \url{https://arxiv.org/abs/2608.05872}.

\bibitem[Belrose et~al.(2023)Belrose, Ostrovsky, McKinney, Furman, Smith,
  Halawi, Biderman, and Steinhardt]{belrose2023tuned}
Nora Belrose, Igor Ostrovsky, Lev McKinney, Zach Furman, Logan Smith, Danny
  Halawi, Stella Biderman, and Jacob Steinhardt.
\newblock {Eliciting Latent Predictions from Transformers with the Tuned Lens},
  2023.
\newblock URL \url{https://arxiv.org/abs/2303.08112}.

\bibitem[Cobbe et~al.(2021)Cobbe, Kosaraju, Bavarian, Chen, Jun, Kaiser,
  Plappert, Tworek, Hilton, Nakano, Hesse, and Schulman]{cobbe2021gsm8k}
Karl Cobbe, Vineet Kosaraju, Mohammad Bavarian, Mark Chen, Heewoo Jun, Lukasz
  Kaiser, Matthias Plappert, Jerry Tworek, Jacob Hilton, Reiichiro Nakano,
  Christopher Hesse, and John Schulman.
\newblock Training verifiers to solve math word problems, 2021.
\newblock URL \url{https://arxiv.org/abs/2110.14168}.

\bibitem[{DeepSeek-AI} et~al.(2025){DeepSeek-AI}, Guo, Yang, Zhang, Song,
  Zhang, Xu, Zhu, Ma, Wang, et~al.]{deepseekai2025r1}
{DeepSeek-AI}, Daya Guo, Dejian Yang, Haowei Zhang, Junxiao Song, Ruoyu Zhang,
  Runxin Xu, Qihao Zhu, Shirong Ma, Peiyi Wang, et~al.
\newblock {DeepSeek-R1}: Incentivizing reasoning capability in {LLMs} via
  reinforcement learning, 2025.
\newblock URL \url{https://arxiv.org/abs/2501.12948v1}.

\bibitem[Dehghani et~al.(2019)Dehghani, Gouws, Vinyals, Uszkoreit, and
  Kaiser]{dehghani2019universal}
Mostafa Dehghani, Stephan Gouws, Oriol Vinyals, Jakob Uszkoreit, and {\L}ukasz
  Kaiser.
\newblock {Universal Transformers}.
\newblock In \emph{International Conference on Learning Representations}, 2019.
\newblock URL \url{https://arxiv.org/abs/1807.03819}.

\bibitem[Furlanello et~al.(2018)Furlanello, Lipton, Tschannen, Itti, and
  Anandkumar]{furlanello2018born}
Tommaso Furlanello, Zachary~C. Lipton, Michael Tschannen, Laurent Itti, and
  Anima Anandkumar.
\newblock {Born Again Neural Networks}.
\newblock In \emph{International Conference on Machine Learning}, 2018.
\newblock URL \url{https://proceedings.mlr.press/v80/furlanello18a.html}.

\bibitem[Geiping et~al.(2025)Geiping, McLeish, Jain, Kirchenbauer, Singh,
  Bartoldson, Kailkhura, Bhatele, and Goldstein]{geiping2025recurrent}
Jonas Geiping, Sean McLeish, Neel Jain, John Kirchenbauer, Siddharth Singh,
  Brian~R. Bartoldson, Bhavya Kailkhura, Abhinav Bhatele, and Tom Goldstein.
\newblock {Scaling up Test-Time Compute with Latent Reasoning: A Recurrent
  Depth Approach}, 2025.
\newblock URL \url{https://arxiv.org/abs/2502.05171}.

\bibitem[Gu et~al.(2024)Gu, Dong, Wei, and Huang]{gu2024minillm}
Yuxian Gu, Li~Dong, Furu Wei, and Minlie Huang.
\newblock {MiniLLM}: Knowledge distillation of large language models.
\newblock In \emph{International Conference on Learning Representations}, 2024.
\newblock URL
  \url{https://proceedings.iclr.cc/paper_files/paper/2024/hash/8ac015d409635f196f9e3e9dcfb9a94e-Abstract-Conference.html}.

\bibitem[Hendrycks et~al.(2021)Hendrycks, Burns, Kadavath, Arora, Basart, Tang,
  Song, and Steinhardt]{hendrycks2021math}
Dan Hendrycks, Collin Burns, Saurav Kadavath, Akul Arora, Steven Basart, Eric
  Tang, Dawn Song, and Jacob Steinhardt.
\newblock {Measuring Mathematical Problem Solving With the MATH Dataset}, 2021.
\newblock URL \url{https://arxiv.org/abs/2103.03874}.

\bibitem[Hinton et~al.(2015)Hinton, Vinyals, and Dean]{hinton2015distilling}
Geoffrey Hinton, Oriol Vinyals, and Jeff Dean.
\newblock {Distilling the Knowledge in a Neural Network}, 2015.
\newblock URL \url{https://arxiv.org/abs/1503.02531}.

\bibitem[Hou et~al.(2025)Hou, Zhang, Ji, Liu, Qian, Andreas, and
  Chang]{hou2025thinkprune}
Bairu Hou, Yang Zhang, Jiabao Ji, Yujian Liu, Kaizhi Qian, Jacob Andreas, and
  Shiyu Chang.
\newblock {ThinkPrune: Pruning Long Chain-of-Thought of LLMs via Reinforcement
  Learning}, 2025.
\newblock URL \url{https://arxiv.org/abs/2504.01296}.

\bibitem[Li et~al.(2024)Li, Cui, Zhao, Kong, and Bi]{li2024gsmplus}
Qintong Li, Leyang Cui, Xueliang Zhao, Lingpeng Kong, and Wei Bi.
\newblock {GSM}-plus: A comprehensive benchmark for evaluating the robustness
  of {LLMs} as mathematical problem solvers.
\newblock In \emph{Proceedings of the 62nd Annual Meeting of the Association
  for Computational Linguistics (Volume 1: Long Papers)}, pp.\  2961--2984.
  Association for Computational Linguistics, 2024.
\newblock \doi{10.18653/v1/2024.acl-long.163}.
\newblock URL \url{https://aclanthology.org/2024.acl-long.163/}.

\bibitem[Li et~al.(2026)Li, Li, and Zhou]{li2026polar}
Ziyue Li, Yang Li, and Tianyi Zhou.
\newblock Skip a layer or loop it? learning program-of-layers in {LLMs}.
\newblock In \emph{International Conference on Machine Learning}, 2026.
\newblock URL \url{https://arxiv.org/abs/2606.06574v2}.

\bibitem[Lightman et~al.(2024)Lightman, Kosaraju, Burda, Edwards, Baker, Lee,
  Leike, Schulman, Sutskever, and Cobbe]{lightman2024verify}
Hunter Lightman, Vineet Kosaraju, Yura Burda, Harri Edwards, Bowen Baker, Teddy
  Lee, Jan Leike, John Schulman, Ilya Sutskever, and Karl Cobbe.
\newblock {Let's Verify Step by Step}.
\newblock In \emph{International Conference on Learning Representations}, 2024.
\newblock URL \url{https://arxiv.org/abs/2305.20050}.

\bibitem[Mozer et~al.(2026)Mozer, Siddiqui, Sawyer, Sanyal, and
  Liu]{mozer2026recirculation}
Michael~C. Mozer, Shoaib~Ahmed Siddiqui, Danny Sawyer, Sunny Sanyal, and
  Rosanne Liu.
\newblock {Recirculation}, 2026.
\newblock URL \url{https://arxiv.org/abs/2608.17981}.

\bibitem[Muennighoff et~al.(2025)Muennighoff, Yang, Shi, Li, Fei-Fei,
  Hajishirzi, Zettlemoyer, Liang, Cand{\`e}s, and Hashimoto]{muennighoff2025s1}
Niklas Muennighoff, Zitong Yang, Weijia Shi, Xiang~Lisa Li, Li~Fei-Fei,
  Hannaneh Hajishirzi, Luke Zettlemoyer, Percy Liang, Emmanuel Cand{\`e}s, and
  Tatsunori Hashimoto.
\newblock {s1: Simple test-time scaling}, 2025.
\newblock URL \url{https://arxiv.org/abs/2501.19393}.

\bibitem[Ng(2026)]{ng2026rys}
David~Noel Ng.
\newblock {LLM} neuroanatomy: How i topped the {LLM} leaderboard without
  changing a single weight.
\newblock Author's technical blog, 2026.
\newblock URL \url{https://dnhkng.github.io/posts/rys/}.

\bibitem[{nostalgebraist}(2020)]{nostalgebraist2020logitlens}
{nostalgebraist}.
\newblock Interpreting {GPT}: The logit lens.
\newblock LessWrong, 2020.
\newblock URL
  \url{https://www.lesswrong.com/posts/AcKRB8wDpdaN6v6ru/interpreting-gpt-the-logit-lens}.

\bibitem[Patel et~al.(2021)Patel, Bhattamishra, and Goyal]{patel2021svamp}
Arkil Patel, Satwik Bhattamishra, and Navin Goyal.
\newblock Are {NLP} models really able to solve simple math word problems?
\newblock In \emph{Proceedings of the 2021 Conference of the North American
  Chapter of the Association for Computational Linguistics: Human Language
  Technologies}, pp.\  2080--2094. Association for Computational Linguistics,
  2021.
\newblock \doi{10.18653/v1/2021.naacl-main.168}.
\newblock URL \url{https://aclanthology.org/2021.naacl-main.168/}.

\bibitem[Sang et~al.(2026)Sang, Xu, Zhou, He, Wang, and Sun]{sang2026crisp}
Hejian Sang, Yuanda Xu, Zhengze Zhou, Ran He, Zhipeng Wang, and Jiachen Sun.
\newblock {CRISP}: Compressed reasoning via iterative self-policy distillation,
  2026.
\newblock URL \url{https://arxiv.org/abs/2603.05433v7}.

\bibitem[Shao et~al.(2024)Shao, Wang, Zhu, Xu, Song, Bi, Zhang, Zhang, Li, Wu,
  and Guo]{shao2024deepseekmath}
Zhihong Shao, Peiyi Wang, Qihao Zhu, Runxin Xu, Junxiao Song, Xiao Bi, Haowei
  Zhang, Mingchuan Zhang, Y.~K. Li, Y.~Wu, and Daya Guo.
\newblock {DeepSeekMath}: Pushing the limits of mathematical reasoning in open
  language models, 2024.
\newblock URL \url{https://arxiv.org/abs/2402.03300}.

\bibitem[Singh et~al.(2024)Singh, Co-Reyes, Agarwal, Anand, Patil, Garcia, Liu,
  Harrison, Lee, Xu, Parisi, Kumar, Alemi, Rizkowsky, Nova, Adlam, Bohnet,
  Elsayed, Sedghi, Mordatch, Simpson, Gur, Snoek, Pennington, Hron, Kenealy,
  Swersky, Mahajan, Culp, Xiao, Bileschi, Constant, Novak, Liu, Warkentin,
  Qian, Bansal, Dyer, Neyshabur, Sohl-Dickstein, and Fiedel]{singh2024rest}
Avi Singh, John~D. Co-Reyes, Rishabh Agarwal, Ankesh Anand, Piyush Patil,
  Xavier Garcia, Peter~J. Liu, James Harrison, Jaehoon Lee, Kelvin Xu, Aaron
  Parisi, Abhishek Kumar, Alex Alemi, Alex Rizkowsky, Azade Nova, Ben Adlam,
  Bernd Bohnet, Gamaleldin Elsayed, Hanie Sedghi, Igor Mordatch, Isabelle
  Simpson, Izzeddin Gur, Jasper Snoek, Jeffrey Pennington, Jiri Hron, Kathleen
  Kenealy, Kevin Swersky, Kshiteej Mahajan, Laura Culp, Lechao Xiao, Maxwell~L.
  Bileschi, Noah Constant, Roman Novak, Rosanne Liu, Tris Warkentin, Yundi
  Qian, Yamini Bansal, Ethan Dyer, Behnam Neyshabur, Jascha Sohl-Dickstein, and
  Noah Fiedel.
\newblock {Beyond Human Data: Scaling Self-Training for Problem-Solving with
  Language Models}.
\newblock \emph{Transactions on Machine Learning Research}, 2024.
\newblock URL \url{https://arxiv.org/abs/2312.06585}.

\bibitem[Tafjord et~al.(2021)Tafjord, Dalvi, and Clark]{tafjord2021proofwriter}
Oyvind Tafjord, Bhavana Dalvi, and Peter Clark.
\newblock {ProofWriter}: Generating implications, proofs, and abductive
  statements over natural language.
\newblock In \emph{Findings of the Association for Computational Linguistics:
  ACL-IJCNLP 2021}, pp.\  3621--3634. Association for Computational
  Linguistics, 2021.
\newblock \doi{10.18653/v1/2021.findings-acl.317}.
\newblock URL \url{https://aclanthology.org/2021.findings-acl.317/}.

\bibitem[Wei et~al.(2022)Wei, Wang, Schuurmans, Bosma, Ichter, Xia, Chi, Le,
  and Zhou]{wei2022chain}
Jason Wei, Xuezhi Wang, Dale Schuurmans, Maarten Bosma, Brian Ichter, Fei Xia,
  Ed~Chi, Quoc Le, and Denny Zhou.
\newblock {Chain-of-Thought Prompting Elicits Reasoning in Large Language
  Models}.
\newblock In \emph{Advances in Neural Information Processing Systems}, 2022.
\newblock URL \url{https://arxiv.org/abs/2201.11903}.

\bibitem[Xia et~al.(2025)Xia, Leong, Wang, Li, and Li]{xia2025tokenskip}
Heming Xia, Chak~Tou Leong, Wenjie Wang, Yongqi Li, and Wenjie Li.
\newblock {TokenSkip: Controllable Chain-of-Thought Compression in LLMs}.
\newblock In \emph{Proceedings of the 2025 Conference on Empirical Methods in
  Natural Language Processing}, 2025.
\newblock URL \url{https://aclanthology.org/2025.emnlp-main.165/}.

\bibitem[Xu et~al.(2026)Xu, Dong, Wang, Sahoo, Li, and Xiong]{xu2026elastic}
Yuhui Xu, Hanze Dong, Lei Wang, Doyen Sahoo, Junnan Li, and Caiming Xiong.
\newblock Scalable chain of thoughts via elastic reasoning.
\newblock In \emph{International Conference on Learning Representations}, 2026.
\newblock URL \url{https://openreview.net/forum?id=E0Qfhma53J}.

\bibitem[Yang et~al.(2025)Yang, Li, Yang, Zhang, Hui, Zheng, Yu, Gao, Huang,
  Lv, Zheng, Liu, Zhou, Huang, Hu, Ge, Wei, Lin, Tang, Yang, Tu, Zhang, Yang,
  Yang, Zhou, Zhou, Lin, Dang, Bao, Yang, Yu, Deng, Li, Xue, Li, Zhang, Wang,
  Zhu, Men, Gao, Liu, Luo, Li, Tang, Yin, Ren, Wang, Zhang, Ren, Fan, Su,
  Zhang, Zhang, Wan, Liu, Wang, Cui, Zhang, Zhou, and Qiu]{yang2025qwen3}
An~Yang, Anfeng Li, Baosong Yang, Beichen Zhang, Binyuan Hui, Bo~Zheng, Bowen
  Yu, Chang Gao, Chengen Huang, Chenxu Lv, Chujie Zheng, Dayiheng Liu, Fan
  Zhou, Fei Huang, Feng Hu, Hao Ge, Haoran Wei, Huan Lin, Jialong Tang, Jian
  Yang, Jianhong Tu, Jianwei Zhang, Jianxin Yang, Jiaxi Yang, Jing Zhou,
  Jingren Zhou, Junyang Lin, Kai Dang, Keqin Bao, Kexin Yang, Le~Yu, Lianghao
  Deng, Mei Li, Mingfeng Xue, Mingze Li, Pei Zhang, Peng Wang, Qin Zhu, Rui
  Men, Ruize Gao, Shixuan Liu, Shuang Luo, Tianhao Li, Tianyi Tang, Wenbiao
  Yin, Xingzhang Ren, Xinyu Wang, Xinyu Zhang, Xuancheng Ren, Yang Fan, Yang
  Su, Yichang Zhang, Yinger Zhang, Yu~Wan, Yuqiong Liu, Zekun Wang, Zeyu Cui,
  Zhenru Zhang, Zhipeng Zhou, and Zihan Qiu.
\newblock {Qwen3} technical report, 2025.
\newblock URL \url{https://arxiv.org/abs/2505.09388}.

\bibitem[Yi et~al.(2025)Yi, Wang, and Li]{yi2025shorterbetter}
Jingyang Yi, Jiazheng Wang, and Sida Li.
\newblock {ShorterBetter: Guiding Reasoning Models to Find Optimal Inference
  Length for Efficient Reasoning}, 2025.
\newblock URL \url{https://arxiv.org/abs/2504.21370}.

\bibitem[Zelikman et~al.(2022)Zelikman, Wu, Mu, and Goodman]{zelikman2022star}
Eric Zelikman, Yuhuai Wu, Jesse Mu, and Noah~D. Goodman.
\newblock {STaR: Bootstrapping Reasoning With Reasoning}.
\newblock In \emph{Advances in Neural Information Processing Systems}, 2022.
\newblock URL \url{https://arxiv.org/abs/2203.14465}.

\bibitem[Zhao et~al.(2026)Zhao, Xie, Liu, Huang, Pang, Chen, and
  Grover]{zhao2026selfdistilled}
Siyan Zhao, Zhihui Xie, Mengchen Liu, Jing Huang, Guan Pang, Feiyu Chen, and
  Aditya Grover.
\newblock Self-distilled reasoner: On-policy self-distillation for large
  language models, 2026.
\newblock URL \url{https://arxiv.org/abs/2601.18734v3}.

\end{thebibliography}
\bibliographystyle{iclr2027_conference}
\clearpage
\appendix
\section{Implementation Details}
\label{sec:implementation}
\paragraph{Optimization.}
Each distillation round uses 200 optimizer updates, a learning rate of $2\times10^{-6}$, and training seed 42. A batch comprises twelve questions and four responses per question. Responses are generated with a single-stage cap of 1,024 tokens, temperature 1, and top-$p$ 1. Teacher weights and the selected route remain fixed throughout a round. The teacher evaluates full-softmax probabilities of student-sampled tokens; the bounded log-probability signal and importance-weighted surrogate are given in Equation~\ref{eq:sampled-loss}. Labels are used for teacher selection; student distillation uses questions and generated responses.

\paragraph{Execution and continuation.}
The main route family repeats a contiguous interval of pretrained layers. Search and route validation compare the candidates at the configured thinking budget, including the standard path. A new round starts from the updated standard student. Subsequent route-based continuation requires a useful nonstandard route under the run's validation rule, with a maximum of ten rounds. The concise-teacher configuration retains its permitted initial prompt-only fallback.

\paragraph{Generation and scoring.}
Main evaluation is greedy and reserves 128 tokens for the answer stage after the configured thinking budget. A task-specific answer cue starts the answer stage. Accuracy is computed by the task's deterministic local rule on the complete fixed evaluation set. Thinking, answer, and total lengths are measured with the corresponding model tokenizer, averaged over every evaluated example. Source identities for length comparisons bind the token counts and correctness outcomes to the same responses.

\paragraph{Length evaluation.}
The paired Elastic accuracy--length evaluation in Figure~\ref{fig:length} scores 76.27\%; its independent main-table evaluation scores 75.59\%. Each point pairs correctness and token counts from the same responses. Token counts describe generated length, not teacher computation or runtime.

\paragraph{Candidate screening and selection.}
We enumerate contiguous repeated-layer intervals and retain the standard route at every stage. We shuffle the official GSM8K training split with seed 42, assigning the first 1000 questions to search and the next 100 to validation; all questions outside validation form the distillation pool. On GSM8K, the first screening stage evaluates routes on 128 search questions and retains the top 24 nonstandard candidates; the second evaluates the survivors on the remaining 872 search questions and retains the top five. Both stages also retain every nonstandard candidate tied at the cutoff. The surviving candidates and standard route are then evaluated on 100 separate validation questions, using identical prompts and token budgets. If the standard route ties for the best validation score, it is retained. Otherwise, ties among the best nonstandard routes are resolved by sampling from their sorted candidate identifiers with seed 42. Thus, adopting a loop requires a strict validation improvement over the standard route. The concise-teacher configuration's initial prompt-only fallback is described above.

\paragraph{Sampled distillation objective.}
Training uses a sampled-token policy-gradient surrogate. For a sampled token, the teacher--student log-probability difference supplies a bounded update signal:
\begin{align}
 a_t &= \operatorname{clip}\!\left(\log q_k(y_t\mid x,y_{<t})
             -\log p_{\mathrm{old}}(y_t\mid x,y_{<t}),-c,c\right), \\
 \rho_t(\theta)&=\frac{p_{\theta,\mathrm{std}}(y_t\mid x,y_{<t})}
                         {p_{\mathrm{old}}(y_t\mid x,y_{<t})},\\
 \widehat{\mathcal L}(\theta)&=
 -\frac{\sum_{x,y,t}m_t\rho_t(\theta)\sg(a_t)}{\sum_{x,y,t}m_t}.
 \label{eq:sampled-loss}
\end{align}
Here, $m_t$ selects valid response tokens, and $\sg$ stops gradients. The importance ratio $\rho_t$ is not clipped. We obtain each sampled token's probability from the full-vocabulary softmax. The training loss uses these sampled-token scores without explicitly computing the full-vocabulary KL divergence.

A positive $a_t$ increases the likelihood of a sampled token that the teacher favors more than the rollout policy; a negative value reduces it. The clipping threshold is $c=10$. Before clipping, the log-probability difference supplies the reverse-KL score-function direction at the rollout policy.

At each training step, we sample a fresh batch of responses using the current student policy, denoted by $p_{\mathrm{old}}$. Each batch is used for one student update. The updated student weights are then synchronized to the rollout workers before the next batch is generated.

\FloatBarrier

\section{Constructing Stronger Teachers through Layer Search}
\label{sec:teachers}
\label{sec:route-landscapes}
\begin{figure}[!htbp]
\centering
\includegraphics[width=0.88\textwidth]{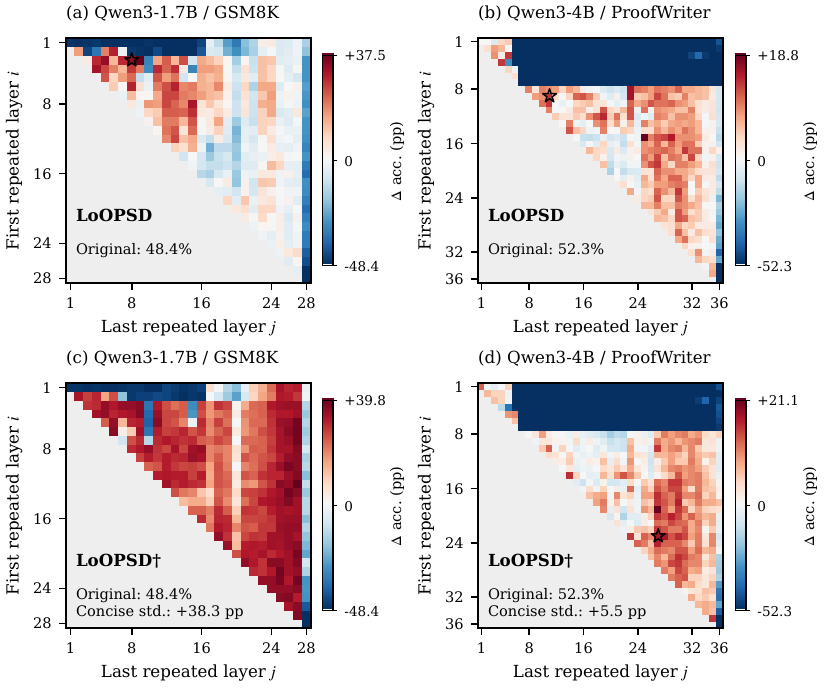}
\caption{Teacher construction on GSM8K with Qwen3-1.7B and ProofWriter with Qwen3-4B, at a 400-token thinking budget. Heatmaps show search accuracy changes in percentage points; each panel has an independent color bar. Stars mark validation-selected nonstandard routes. Validation selects a repeated-layer route under both prompts on ProofWriter, while retaining standard execution under the concise prompt on GSM8K.}
\label{fig:routes-q17}
\end{figure}

\begin{figure}[!htbp]
\centering
\includegraphics[width=0.88\textwidth]{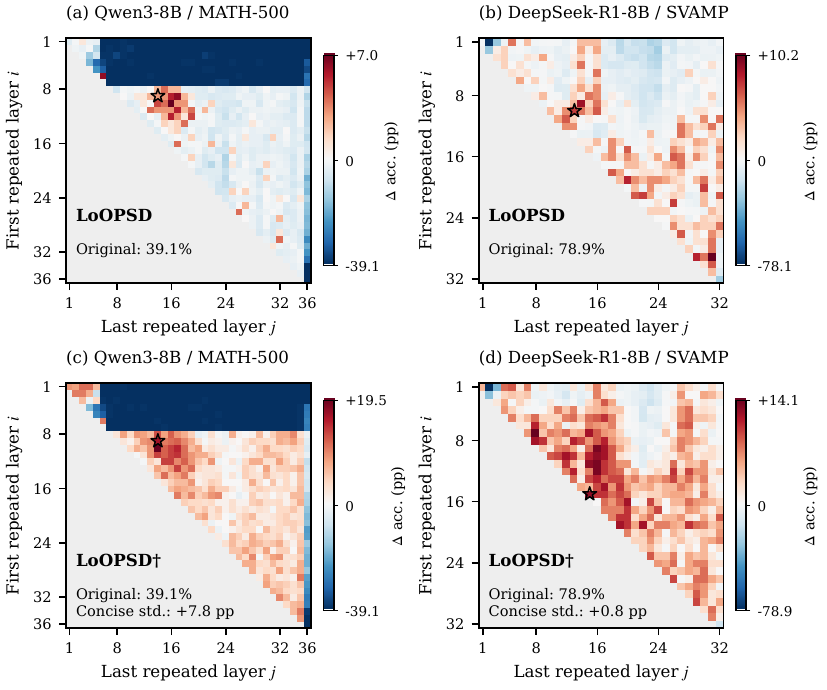}
\caption{Teacher construction on MATH-500 with Qwen3-8B and SVAMP with DeepSeek-R1-8B, using the same 400-token protocol. Heatmaps show search accuracy changes; independent bars preserve each panel's measured range. Stars mark validation-selected nonstandard routes, which are selected under both the original and concise prompts in these two settings.}
\label{fig:routes-q4}
\end{figure}

We examine whether repeating layers can improve the teacher without changing its weights, and whether this source of supervision remains useful with a concise teacher prompt. Figures~\ref{fig:routes-q17} and~\ref{fig:routes-q4} show first-round route-search landscapes across four model--task settings. All settings use 400 thinking tokens and 128 answer tokens. Each landscape evaluates all repeated-layer intervals on the same 128 search questions within its setting; a separate validation set selects the teacher.

The maps show accuracy changes relative to the original-prompt standard model. Concise-standard gains printed inside the lower panels identify the contribution of the prompt before any layers are repeated. Each panel has its own red--white--blue scale, centered at zero and spanning all measured values; negative and positive ranges are scaled separately. Thus, equal color intensity across panels need not indicate equal accuracy changes. In the concise-prompt panels, the color includes both the prompt contribution and any additional change from layer repetition. The axes give the first and last repeated layers in the inclusive interval $[i,j]$. Stars mark nonstandard routes selected by validation; no star is shown when the standard route is selected.

\paragraph{From a route landscape to a stronger teacher.}
Beneficial regions depend on the task and teacher prompt. Under the original prompt, validation selects a nonstandard route in all four displayed settings. Under concise prompting, validation selects nonstandard routes on ProofWriter, MATH-500, and SVAMP, while retaining the standard route on GSM8K. These landscapes show why the repeated interval is searched for each setting, and how the procedure can retain standard execution when repetition provides no validated advantage.

\section{Transfer beyond GSM8K}
\label{sec:math-transfer}
We evaluate whether the improvements learned on GSM8K transfer to SVAMP \citep{patel2021svamp} and GSM-Plus \citep{li2024gsmplus}. We reuse the frozen GSM8K checkpoints from the 400-token comparison, including the shared 200-update CRISP checkpoint. Target-task examples are used only for evaluation; they do not enter route search, training, or checkpoint selection. SVAMP contributes its fixed 300-question test split, while GSM-Plus contributes a fixed 1,000-question subset of \texttt{testmini}. All methods use the original math prompt, standard layer execution, greedy decoding, 400 thinking tokens, and a separate 128-token answer budget.

\begin{table}[!htbp]
\centering
\caption{\textbf{Transfer of GSM8K-trained students to additional math datasets.} Accuracy (\%) and mean thinking tokens at a 400-token thinking cap. All methods use standard execution, the original prompt, and greedy decoding.}
\label{tab:math-transfer}
\small
\setlength{\tabcolsep}{6pt}
\renewcommand{\arraystretch}{1.12}
\begin{tabular}{p{108pt}*{4}{>{\raggedleft\arraybackslash}p{\dimexpr(\linewidth-168pt)/4\relax}}}
\toprule
& \multicolumn{2}{c}{\textbf{SVAMP}} & \multicolumn{2}{c}{\textbf{GSM-Plus}} \\
\cmidrule(lr){2-3}\cmidrule(lr){4-5}
\hspace{5pt}Method & Accuracy & Think tokens & Accuracy & Think tokens \\
\midrule
\rowcolor{modelband}
\multicolumn{5}{l}{\rule{0pt}{12.5pt}\textbf{Qwen3-1.7B}\rule[-3pt]{0pt}{0pt}} \\
\hspace{5pt}Original & 71.33 & 299.7 & 31.70 & 378.8 \\
\hspace{5pt}Elastic Reasoning & 75.00 & 294.4 & 30.20 & 378.1 \\
\hspace{5pt}CRISP & 87.67 & 251.8 & 58.50 & 353.5 \\
\rowcolor{methodband}
\hspace{5pt}LoOPSD & 85.33 & 373.0 & 61.40 & 392.7 \\
\rowcolor{methodband}
\hspace{5pt}LoOPSD$^{\dagger}$ & \textbf{90.33} & 217.6 & \textbf{61.80} & 327.7 \\
\specialrule{0.3pt}{3pt}{3pt}
\rowcolor{modelband}
\multicolumn{5}{l}{\rule{0pt}{12.5pt}\textbf{Qwen3-4B}\rule[-3pt]{0pt}{0pt}} \\
\hspace{5pt}Original & 87.33 & 321.1 & 56.40 & 385.3 \\
\hspace{5pt}Elastic Reasoning & 88.33 & 316.6 & 58.90 & 384.6 \\
\hspace{5pt}CRISP & 92.00 & 258.3 & 70.10 & 343.0 \\
\rowcolor{methodband}
\hspace{5pt}LoOPSD & 92.67 & 241.2 & \textbf{74.20} & 332.7 \\
\rowcolor{methodband}
\hspace{5pt}LoOPSD$^{\dagger}$ & \textbf{93.67} & 201.5 & 73.20 & 295.9 \\
\specialrule{0.3pt}{3pt}{3pt}
\rowcolor{modelband}
\multicolumn{5}{l}{\rule{0pt}{12.5pt}\textbf{Qwen3-8B}\rule[-3pt]{0pt}{0pt}} \\
\hspace{5pt}Original & 89.33 & 354.1 & 61.00 & 390.9 \\
\hspace{5pt}Elastic Reasoning & 90.00 & 351.5 & 61.20 & 390.1 \\
\hspace{5pt}CRISP & \textbf{93.67} & 277.1 & 70.00 & 347.0 \\
\rowcolor{methodband}
\hspace{5pt}LoOPSD & 91.33 & 284.9 & 74.20 & 343.7 \\
\rowcolor{methodband}
\hspace{5pt}LoOPSD$^{\dagger}$ & 92.67 & 251.7 & \textbf{74.30} & 320.3 \\
\specialrule{0.3pt}{3pt}{3pt}
\rowcolor{modelband}
\multicolumn{5}{l}{\rule{0pt}{12.5pt}\textbf{DeepSeek-R1-Distill-Llama-8B}\rule[-3pt]{0pt}{0pt}} \\
\hspace{5pt}Original & 76.67 & 116.4 & 49.00 & 197.3 \\
\hspace{5pt}Elastic Reasoning & 81.33 & 275.7 & 53.80 & 340.8 \\
\hspace{5pt}CRISP & 78.67 & 135.3 & 52.80 & 253.9 \\
\rowcolor{methodband}
\hspace{5pt}LoOPSD & 84.00 & 340.5 & 57.60 & 380.3 \\
\rowcolor{methodband}
\hspace{5pt}LoOPSD$^{\dagger}$ & \textbf{87.33} & 286.6 & \textbf{59.50} & 357.3 \\
\bottomrule
\end{tabular}
\par\vspace{4pt}
\begin{minipage}{\linewidth}\footnotesize
\textbf{Bold}: highest accuracy per model and dataset. $\dagger$: concise teacher prompt during distillation. SVAMP uses the fixed 300-question test split; GSM-Plus uses the fixed 1,000-question evaluation subset. Answer cap: 128 tokens. Thinking lengths are tokenizer-specific averages over every evaluated example.
\end{minipage}
\end{table}

LoOPSD and \lpd{} improve over the original model on both target datasets for all four models. On GSM-Plus, ordinary LoOPSD also exceeds CRISP in every model, by 2.9--4.8 percentage points. The concise-teacher configuration uses fewer thinking tokens than ordinary LoOPSD in all eight model--dataset comparisons. These results show that the learned gains extend beyond the training benchmark to additional math problems and GSM8K perturbations.
\FloatBarrier

\section{From Initial Teachers to Final Students}
\label{sec:extended}
\begin{figure}[p]
\centering
\includegraphics[width=0.90\textwidth]{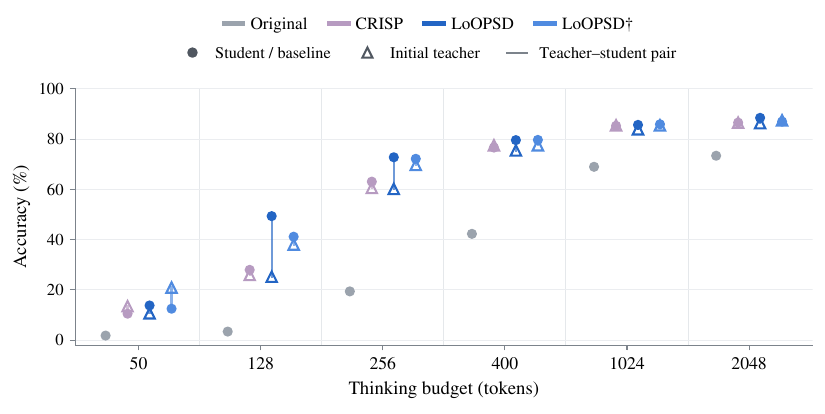}
\caption{Initial teachers and final students on GSM8K with Qwen3-1.7B. Each initial teacher is connected to the final student obtained from the corresponding training configuration. Final LoOPSD students outperform their initial teachers at all six thinking budgets.}
\label{fig:extended0}
\vspace{1em}
\includegraphics[width=0.90\textwidth]{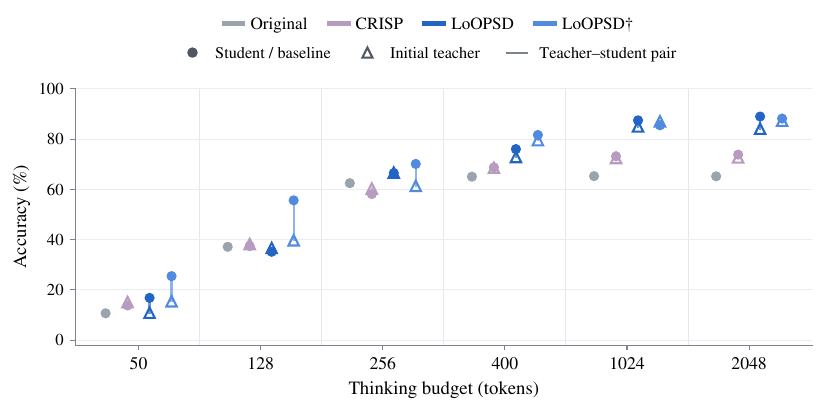}
\caption{Initial teachers and final students on GSM8K with DeepSeek-R1-Distill-Llama-8B. Each initial teacher is connected to the final student obtained from the corresponding training configuration. Final LoOPSD students match or exceed their initial teachers at most thinking budgets.}
\label{fig:extended2}
\end{figure}

We ask whether the final standard-path student catches up to the initial looped teacher found before distillation. Figures~\ref{fig:extended0} and~\ref{fig:extended2} compare the original model, the initial teacher, and the final student across six thinking budgets on GSM8K. Each budget uses its corresponding search and training configuration. Because LoOPSD may continue for multiple rounds, this comparison does not measure one-round transfer; instead, it shows where the final student ends up relative to the teacher that started the training process.

Filled circles denote students or original models, open triangles denote initial teachers, and vertical segments connect each teacher with its final student.

For Qwen3-1.7B, the final LoOPSD student outperforms its initial teacher at all six budgets. For DeepSeek-R1-Distill-Llama-8B, the final student matches or exceeds the initial teacher in most settings. These results show that iterative LoOPSD can move the standard-path student beyond the performance of the looped teacher available at the start of training.

\FloatBarrier

\clearpage
\section{Layerwise Gold-Label Log Probability}
\label{sec:layerwise-logprob}
\begin{figure}[!htbp]
\centering
\includegraphics[width=0.72\textwidth]{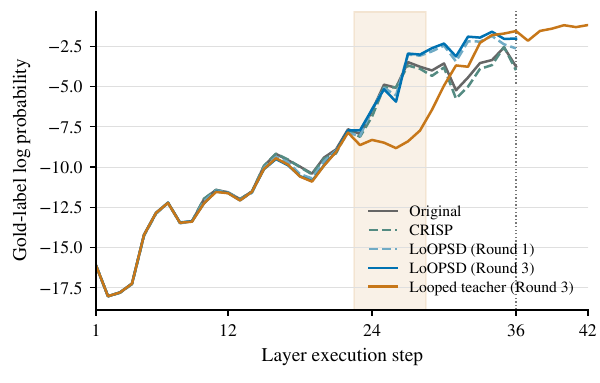}
\caption{Gold-label log probability across layer executions on ProofWriter with Qwen3-4B. Higher values indicate greater probability assigned to the correct label. Shading marks repeated teacher execution. LoOPSD students assign higher probability to the correct label in later layers.}
\label{fig:layerwise-logprob}
\end{figure}

We test whether the stronger intermediate-layer answer readout also reflects greater probability assigned to the correct answer, rather than only changes in the top-ranked label. Figure~\ref{fig:layerwise-logprob} measures gold-label log probability on the same fixed ProofWriter contexts and with the same readout as Figure~\ref{fig:readout}. The fixed readout follows the logit-lens approach \citep{nostalgebraist2020logitlens} and fits no layer-specific probes, unlike the tuned lens \citep{belrose2023tuned}. The original model, CRISP student, first- and third-round LoOPSD students, and third-round looped teacher share inputs, answer positions, and the original model's final normalization and output projection. The later-layer increases in gold-label log probability complement the accuracy gains in Figure~\ref{fig:readout}, showing stronger readout of the correct answer even when its rank does not change.

\section{Sensitivity to Sampling Temperature}
\label{app:temperature}
We evaluate whether the improvements persist under stochastic thinking generation. For each method, we keep the student checkpoint fixed and vary only the thinking temperature; the answer stage uses greedy decoding. Table~\ref{tab:temperature} shows that both LoOPSD variants remain above the original model across all tested temperatures. LoOPSD$^\dagger$ achieves the highest accuracy in seven of the eight model--temperature settings, with LoOPSD leading in the remaining setting.
\begin{table}[ht]
\centering
\small
\setlength{\tabcolsep}{4pt}
\caption{GSM8K accuracy (\%) at different thinking temperatures. We evaluate the same checkpoints with a 400-token thinking budget and a 128-token answer budget, sampling one response per question. LoOPSD improvements persist across decoding temperatures.}
\label{tab:temperature}
\begin{tabular}{lrrrrrrrr}
\toprule
 & \multicolumn{4}{c}{Qwen3-1.7B} & \multicolumn{4}{c}{Qwen3-4B} \\
\cmidrule(lr){2-5}\cmidrule(lr){6-9}
Method & $0$ & $0.3$ & $0.5$ & $1.0$ & $0$ & $0.3$ & $0.5$ & $1.0$ \\
\midrule
Original & 42.38 & 42.99 & 41.55 & 41.70 & 74.91 & 73.62 & 75.89 & 73.92 \\
Elastic & 44.43 & 44.12 & 44.20 & 43.59 & 76.27 & 75.66 & 76.12 & 75.36 \\
CRISP & 76.65 & 78.17 & 79.23 & 77.26 & 87.64 & 88.40 & 88.02 & 87.57 \\
\rowcolor{gray!8}
LoOPSD & 79.61 & 79.45 & 79.38 & 77.03 & 89.69 & 90.37 & 90.37 & \textbf{91.21} \\
\rowcolor{gray!8}
LoOPSD$^\dagger$ & \textbf{79.68} & \textbf{80.52} & \textbf{81.88} & \textbf{80.82} & \textbf{91.05} & \textbf{91.51} & \textbf{91.66} & 91.05 \\
\bottomrule
\end{tabular}
\end{table}

\end{document}